\documentclass[10pt,twocolumn,letterpaper]{article}
\usepackage{cvpr}
\usepackage{times}
\usepackage{epsfig}
\usepackage{graphicx}
\usepackage{amsmath}
\usepackage{amssymb}
\usepackage{nopageno}
\usepackage[pagebackref=true,breaklinks=true,letterpaper=true,colorlinks,bookmarks=false]{hyperref}

\cvprfinalcopy % *** Uncomment this line for the final submission

\def\cvprPaperID{29} % *** Enter the CVPR Paper ID here

\ifcvprfinal\fi
\begin{document}
%%%%%%%%% TITLE
\title{A Deep Face Identification Network Enhanced by Facial Attributes Prediction}

\author{Fariborz Taherkhani,  Nasser M. Nasrabadi,  Jeremy Dawson \\
West Virginia University\\
{\tt\small ft0009@mix.wvu.edu,  nasser.nasrabadi@mail.wvu.edu, Jeremy.Dawson@mail.wvu.edu }
}

\maketitle
\begin{abstract}
  In this paper, we propose a new deep framework which predicts facial attributes and leverage it as a soft modality to improve face identification performance. Our model is an end to end framework which consists of a convolutional neural network (CNN) whose output is fanned out into two separate branches; the first branch predicts facial attributes while the second branch identifies face images. Contrary to the existing multi-task methods which only use a shared CNN feature space to train these two tasks jointly, we fuse the predicted attributes with the features from the face modality in order to improve the face identification performance. Experimental results show that our model brings benefits to both face identification as well as facial attribute prediction performance, especially in the case of identity facial attributes such as gender prediction. We tested our model on two standard datasets annotated by identities and face attributes. Experimental results indicate that the proposed model outperforms most of the current existing face identification and attribute prediction methods.
\end{abstract}

%%%%%%%%% BODY TEXT
\section{Introduction}
Deep neural networks, particularly deep Convolutional Neural Networks (CNNs), have provided significant improvement in visual tasks such as face recognition, attribute prediction and image classification \cite{krizhevsky2012imagenet,simonyan2014very,szegedy2015going,he2016deep,liu2015deep,parkhi2015deep}. Despite this advancement, designing a deep model to learn different tasks jointly while improving their performance by sharing learned parameters remains a challenging problem. 

Providing auxiliary information  to a CNN-based face recognition model can improve its recognition performance; however, in some cases such information is available only during training and may not be available during the testing phase. Despite the potential advantages of using auxiliary data, these problems have diminished the popularity and flexibility of using both soft and hard modalities for biometric applications \cite{talreja2017multibiometric}.

We propose a model which jointly predicts facial attributes and identifies faces while simultaneously leverages the predicted facial attributes as an auxiliary modality to improve face identification performance. We also show that when our model is trained jointly to recognize face images and predict facial attributes, the model performance on facial attribute prediction increases as well. In other words, in our model the two modalities improve each other's performance once they are trained jointly. We show that some soft biometric information, such as age and gender which on their own are not distinctive enough for face identification, but, nevertheless provide complementary information along with other primary information, such as the face images.

Despite significant improvements in face recognition performance, it is still an ongoing problem in computer vision \cite{best2014unconstrained,guillaumin2009you,schroff2015facenet,schwartz2010robust,sun2015deepid3,taigman2015web}. There are a number of approaches in the literature that use facial attributes for biometrics applications such as face recognition. For example, Wang et al \cite{wang2017multi} propose an attribute-constrained face
recognition model for joint facial attributes prediction and face recognition. In this model, the parameters of the network are first updated for attributes prediction and then same network is fine-tuned  for face recognition.  While Ranjan et al \cite{ranjan2017all} add other face related tasks to improve overall performance. Their model is a single multi-task CNN network for simultaneous face detection, face alignment, pose estimation, gender recognition, smile detection, age estimation and face recognition.
 
Facial attributes as semantic features can be predicted from face images directly, or from other facial attributes indirectly \cite{torfason2016face}. Attribute prediction methods are generally classified into local or global approaches. Local methods consist of three steps; first they detect different parts of the object and then extract features from each part. Finally, these features are concatenated to train a classifier \cite{kumar2009attribute,bourdev2011describing,chung2012deep,berg2013poof,luo2013deep,zhang2014panda}. For example, Kumar et al's method \cite{kumar2009attribute} is based on extracting hand-crafted
 features from ten facial parts. Zhang et al \cite{zhang2014panda} extract poselets aligning face parts to predict facial attributes.  This method works improperly if object localization and alignment are not perfect. Global approaches, however, extract features from entire image disregarding object parts and then train a classifier on extracted features; these methods perform improperly if large face variations such as occlusion, pose and lighting
are present in the image \cite{liu2015deep,kalayeh2017improving,luo2013deep}.

 Attribute prediction has been improved in recent years. Bourdev et al \cite{bourdev2016pose} propose a part-based attribute prediction method which deploys semantic segmentation in order to transfer localization information from the auxiliary task of semantic face parsing to the facial attribute prediction task. Liu et al \cite{liu2015deep} use two cascaded CNNs; the first of which, LNet, is used for face localization, while the second, ANet, is used for attribute description. Zhong et al \cite{zhong2016face} first localize face images and then use an off-the-shelf architecture designed for face recognition to describe face attributes at different levels of a CNN. He et al \cite{BMVC2016_131} propose a multi-task framework for relative attribute prediction. The method uses a CNN to learn local context and global style information from the intermediate convolution and fully connected layers, respectively. 
 
 Our network is inspired by multi-task network but we fuse the output of the attribute predictor into the face recognition layers which makes it different from other existing  multi-task methods such as Wang et al's \cite{wang2017multi} approach. Our deep CNN model is constructed from two cascaded networks in which the final one consists of two branches, each of which are used for facial attribute prediction and face identification, respectively. Both these two branches communicate information together by sharing parameters of the  first network in the model as well as  fusing attribute branch with the last pooling layer of the face identification branch. In our model, all the parameters (i.e. the parameters of the two cascaded networks) are updated simultaneously in each training step.
 
 The Contributions of our work are summarized as follows: 

1) We design a new end to end CNN architecture that learns to predict facial attributes while simultaneously being trained with the objective of face identification. Our model shares learned parameters to train both tasks and also  fuses attribute information  and the face modality  to improve face identification performance.

2) Contrary  to  the  existing  multi-task  methods  that only  use  a  shared  CNN  feature  space  to  train  these  two tasks jointly, our model uses a feature level fusion approach to leverage facial attributes for improving face identification performance.  Furthermore, we observe that our jointly trained network  is a more capable face attribute predictor than one trained on facial attributes alone.

 The rest of this paper is organized as follows: The CNN architecture is described in section 2, fusion of attribute and face modalities is described in section 3, model training parameters are described in section 4, and finally, results and concluding remarks are provided in  sections 5 and 6,  respectively.
\begin{figure*}[ht]
\centering
\includegraphics[scale=0.265]{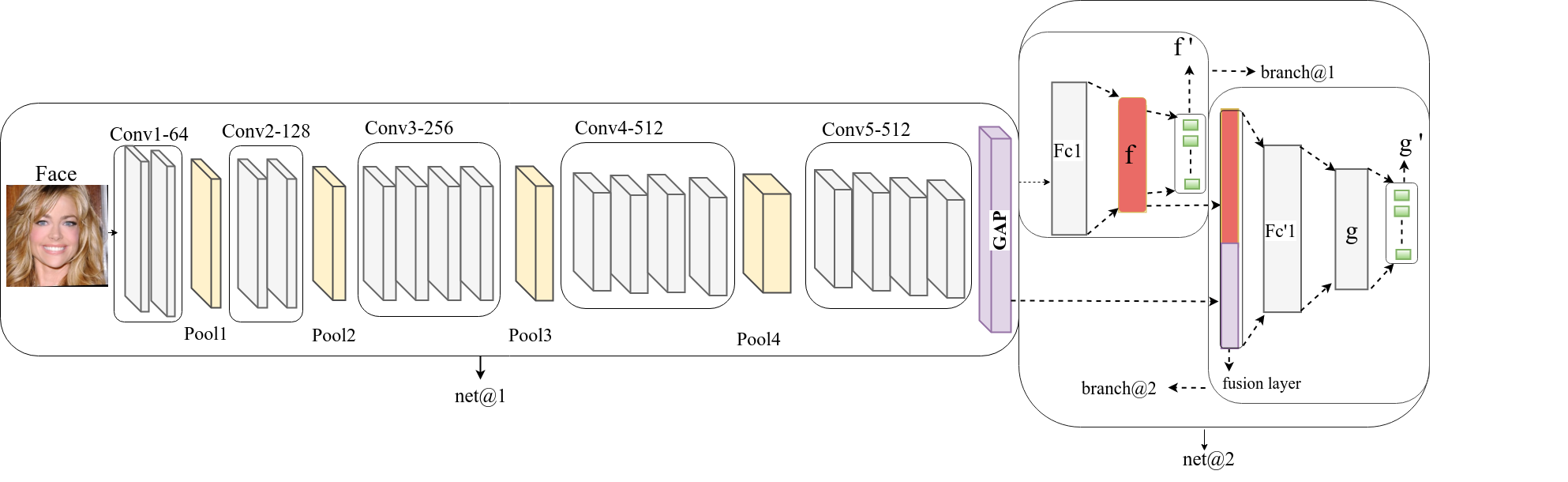}
\caption{Proposed CNN architecture, face identification and attribute prediction are trained jointly.}
\label{fig:arc}
\end{figure*}

\section{Deep Joint Facial Attributes Prediction and Face Identification Model}

The proposed architecture predicts facial attributes and uses them as an auxiliary modality to recognize face images. The model is constructed from two successive cascaded networks as shown in Fig.1. The first network (net@1) uses the VGG 19  structure \cite{simonyan2014very} with identical filter size, convolutional layers, and pooling operation. The first network applies filters with  $3 \times 3$ receptive field.  The convolution stride is set to 1 pixel. To preserve spatial resolution after convolution, spatial padding of the convolutional layer is fixed to 1 pixel for all $3 \times 3$ convolutional layers. Spatial pooling is performed by four max-pooling layers placed after  the second, fourth, eighth, and twelfth convolutional layers and one global average pooling (GAP) layer  which is placed after the sixteenth convolutional layer. Max-pooling is carried out on a $2 \times 2$ pixel window with a stride of 2. Each hidden layer is followed by a Rectified Linear Units (ReLU) \cite{krizhevsky2012imagenet} activation function. A GAP layer is  a substantial process in our model because by disregarding the GAP layer and replacing it by a max-pooling layer, the output of the fusion layer will have a very high dimension when we fuse face and attribute modalities together. The GAP layer simply takes average of each feature map obtained from last convolutional layer. Since no parameter is optimized at the GAP layer, overfitting is prevented at this layer.

The second network (net@2) is divided into two separate branches trained simultaneously while communicating information together through the training process. Both of these  branches consist of two fully connected (FC) layers operating on the output of the first network. The first FC layer of each branch (Fc1 and Fc$^ \prime1$ in Fig.1) consists of 4096 units. The next layers of (Fc1) and (Fc$^ \prime1$) are fully connected layers on which the soft-max operation is conducted. The first branch performs the attribute prediction task, and the output of the last FC layer in this branch before performing soft-max operation is fused with  the GAP layer of net @1 by using Kronecker product \cite{graham1981kronecker}. Finally this fused layer is employed to train the second branch - the face identification task. As shown in Fig.\ref{fig:arc}, attributes are predicted by net@1  and first branch of net@2 parameters while face images are identified by net@1 and all parameters in net@2;
the overall proposed architecture is shown in Fig.\ref{fig:arc}.
\section{ Fusion Layer on Facial Attributes and Face Modalities}

Previously, feature concatenation has been used as an approach for multimodal fusion. In this work, we use the Kronecker product to fuse facial attributes features with face features. Since  the Kronecker product  of two vectors (i.e. attributes and face features) is mathematically formed by a matrix direct product, there are  no learnable parameters at this layer and, consequently, chances of overfitting are low at this layer. Furthermore, we argue that,  due to existing correlation between facial attributes features  and face features,  the output neurons of the fusion layer are simple to interpret and are semantically meaningful. (i.e., the manifold that they will lie on is not complex, however, it is just high dimensional). Therefore, it is simple for the following layers of the network to decode such meaningful information. Assume that \textbf{v} and \textbf{u} are the feature vectors of attributes and face, respectively. The  Kronecker product of these two vectors is defined as follows: 
\begin{equation}
  \textbf{u} \otimes \textbf{v} =
  \begin{bmatrix}
           u_{1} \\
           u_{2} \\
           \vdots \\
           u_{n}\\
         \end{bmatrix}
          {\otimes}
          \begin{bmatrix}
         v_{1} \\
           v_{2} \\
           \vdots \\
           v_{m}
         \end{bmatrix}\\
         =
         \begin{bmatrix}
           u_{1}v_{1} \\
           u_{1}v_{2} \\
           \vdots \\
           u_{1}v_{m}\\
           u_{2}v_{1}\\
            \vdots \\
            u_{n}v_{m}
           \end{bmatrix}
  \end{equation}
\section{Training our CNN architecture} 

In this section, we describe how we train our model.  Thousands of images are needed to train such a deep model. For this reason, we initialize net@1 parameters by a CNN pre-trained on the ImageNet dataset and then we fine tune it as a classifier by using the CASIA-Web Face dataset. CASIA-Web Face contains 10,575 subjects and 494,414 images. As far as we know, this is the largest publicly available face image dataset, second only to the private Facebook dataset.  

The proposed deep network is described as a succession of two cascaded networks. net@1 is constructed from 16 layers of convolutional operations on the inputs, intertwined with ReLU non-linear operation and five pooling operations. Weights in each convolutional layer form a sequence of \textit{4-d} tensors; $ \textit{W}  \in {\rm I\!R}^{l \times c \times p \times q} $ where \textit{l},  \textit{c}, \textit{p} and \textit{q} are  dimensions of the weights along the axes of filter, channel, and spatial width and height, respectively. For notational simplicity, we denote all the weights in net@1 with $W_1$ and the weights in net@2 with $W_2$. $W_2$ is separated into two groups of $W_{2,1}$ and $W_{2,2}$ representing all weights in the first and second branches, respectively. 

$\mathcal{L}_1$ and $\mathcal{L}_2$ described in (2) and (3) are the loss functions designed to perform attribute prediction and face identification tasks, respectively.  We use the cross entropy as our network loss functions. \textit{T, C} and $\textbf{X}=\{x_i\}_{i=1}^N$ indicate the number of facial attributes used in the model,  number of classes and the training samples, respectively. $L_i'$ and $L_{ji}$ represent face label and facial attribute label for attribute \textit{j} and the training sample \textit{i}, respectively. $f$ and $g$ functions are  outputs of the network for attribute prediction and face identification tasks, respectively.  $f'$ and $g'$ are  soft-max functions performed on the $f$ and $g$ outputs, respectively.   The loss functions represented in (2) and (3) show  how two branches of net@2 communicate information and update their learning parameters with each other. As shown in (2) and (3),  the $f$  function (attribute prediction output) takes $W_1$ and $W_{2,1}$  as input. The $g$  function (face identification output) takes $W_1$, $W_{2,2}$ and $f$ as input.  Therefore, both attribute prediction and face identification use $W_1$ as shared parameters.  Furthermore, attribute prediction parameters and $W_{2,2}$ are used for face identification.

We use an Adam optimizer \cite{kingma2014adam} to minimize our network's loss functions. The Adam optimizer is a robust and well-adapted optimizer that can be applied to a variety of non-convex optimization problems in the field of deep neural networks. All parameter values used in Adam optimizer are initialized using the authors' suggestion; we set learning rate to 0.001 to minimize our network's loss functions. 

 The optimization algorithm mainly consists of two steps, the first of which calculates the gradient of the loss functions with respect to the model parameters, and then, for the second step, updates the biased first moment estimate and the model parameters, successively.
\begin{equation}
\begin{split}
 & \mathcal{L}_1(W_1, W_{2,1},X) = -\sum\limits_{j=1}^{T} \sum\limits_{i=1}^{N} L_{ji} log(f'(f( L_{ji}|x_i,W_1,\\ & W_{2,1})))  +(1-L_{ji})log(f'(f(1-L_{ji}|x_i,W_1,W_{2,1})))
\end{split}
\end{equation}

\begin{equation}
\begin{split}
&\mathcal{L}_2(W_1, W_{2,1},W_{2,2},X) =  - \sum\limits_{i=1}^{N} \sum\limits_{k=1}^{C} L'_{ik}\   log(g'(g(L'_{ik}|x_i,\\ & W_1, W_{2,2}, f(x_i,W_1,W_{2,1}))))
\end{split}
\end{equation}

We iterate this algorithm through several epochs for the complete training batches until training error convergence is achieved.
\section{Experiment}
 We conducted experiments for two different cases to examine if our model improves overall performance in identification and prediction tasks. In the first case, we train and test the model to perform two tasks separately in  isolation, while in the second case we employ our model to train both tasks jointly. In the second case, however, we predict facial attributes assuming that such information is not available during the testing phase, and then outputs of the attribute prediction branch before performing the soft-max operation  is fused  with the last pooling layer of net@1 by using the Kronecker product. We fuse the face modality with those facial attributes such as gender and face shape which remain the same in all images of a person. Experimental results show that our model increases overall performance in face identification as well as attribute prediction in comparison to the first case. We performed our experiments on two GeForce GTX TITAN X 12GB GPU.  We ran our model through 100 epochs using batch normalization (i.e. shifting inputs to zero-mean and unit variance)  after each convolutional and fully connected layer before performing no-linearity. Batch normalization potentially helps to achieve faster learning as well as higher overall accuracy. Furthermore, batch normalization allows us to use a higher learning rate, which potentially provides another boost in speed. We used TensorFlow to implement our network. The batch size in all  experiments  is fixed to 128.
 \begin{figure}
\includegraphics[scale=0.35]{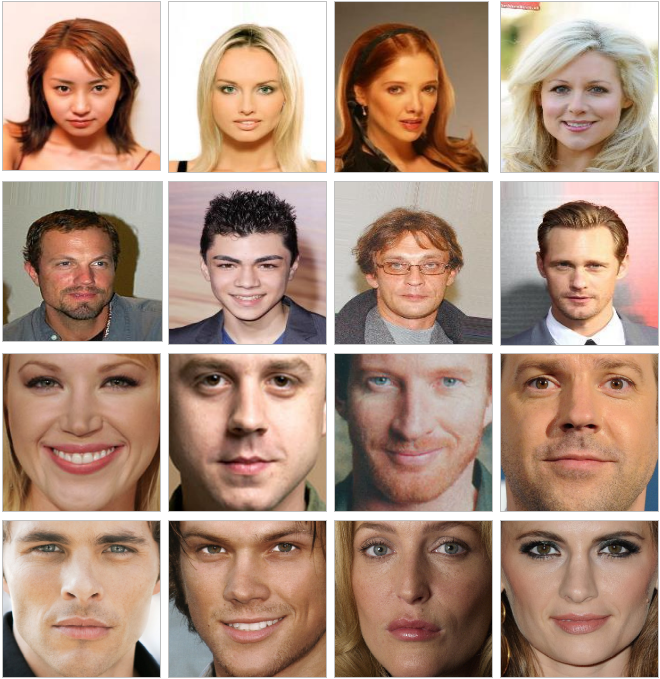}
\begin{center}
\caption{: First and second rows are image samples in CelebA dataset; third and forth rows are samples of aligned face images in MegaFace dataset.}
\end{center}
\end{figure}
 
\subsection{Datasets}

We conducted our experiments on the  CelebA dataset \cite{liu2015deep} for facial attribute prediction, as well as MegaFace \cite{kemelmacher2016megaface} which is a widely used and well-known face datasets for face identification.

\textbf{CelebA} is a large-scale, richly annotated face attribute dataset containing more than 200K celebrity images, each of which is notated with 40 facial attributes. CelebA has about ten thousand identities with twenty images per identity on average. This dataset is also annotated by five landmarks. The dataset can be used  as the training and testing sets for facial attribute prediction, face detection, and landmark (or facial part) localization. To compare our method fairly with the other methods, we use the same setup that they have used. We use images of 8000 identities for training and remaining 1000 identities for testing. Train and test sets are available here.\footnote{http://mmlab.ie.cuhk.edu.hk/projects/CelebA.html}

\textbf{MegaFace} is a publicly available and very challenging dataset which is used for evaluating the performance of  face recognition algorithms with up to a million distractors ( i.e., up to a million people who are not in the test set). MegaFace contains 1M images from 690K individuals with unconstrained pose, expression, lighting, and exposure. MegaFace captures many different subjects rather than many images of a small number of subjects. The gallery set of MegaFace is  collected from a subset of Flickr \cite{thomee2015new}.  The probe set of MegaFace 
used in the challenge consists of two databases; Facescrub \cite{ng2014data}  and FGNet \cite{fu2014interestingness}.  FG-NET contains 975 images of 82 individuals, each with several images spanning ages from 0 to 69. Facescrub dataset contains more than 100K face images of 530 people. The MegaFace challenge evaluates performance of face recognition algorithms by increasing the numbers of “distractors” (going from 10 to 1M) in the gallery set. Training size is important, since it has been shown that face recognition algorithms that were trained on larger sets tend to perform better at scale.
In order to evaluate the face recognition algorithms fairly, MegaFace challenge has two protocols including large or small training sets.
If a training set has more than 0.5M images and 20K subjects,  it is considered as large.  Otherwise, it is considered as small.
We use a small training set which has 0.44M images and 10k subjects. The prob set in our experiments is Facescrub.
%\begin{figure*}[th!]
%\includegraphics[trim={0cm 0 0 0},scale=0.4]{FinalClebLoss.eps}
%\begin{center}
%\caption{ FRTL (left) and APTL (right)  minimization trends for the LFW dataset using AMO. %The figure shows that two tasks (i.e., face recognition and attribute prediction) are %trained jointly without one adversely affecting the other training loss convergence.}
%\end{center}
%\end{figure*}
\subsection{Evaluation metrics} We evaluate the face identification performance of our model on the MegaFace dataset; and facial attribute prediction performance on the CelebA dataset. The MegaFace dataset is not annotated by facial attribute. Our model, however, predicts facial attributes and then uses them for face identification.  To conduct experiments on  the  MegaFace dataset,  we  restore the model parameters trained on  the CelebA dataset, which is annotated by facial attributes as well as people identification, and then fine-tune the model parameters on the MegaFace dataset for the objective of face identification on the MegaFace dataset. Our model predicts facial attributes from the first branch of our architecture and employs this auxiliary modality for face identification.

\textbf{Face Identification:} we calculate the similarity between each of the images in the gallery set and given image from the probe set, and then rank these images based on the obtained similarities. In face identification, the gallery  set should contain at least one image of the same identity. We evaluate our model by using rank-1 identification accuracy as well as Cumulative Match Characteristics (CMC) curves. CMC is a rank-base metric indicating the  probability of the correct gallery image that can be found in the top \textit{k} similar images from the gallery set.
\begin{table*}
\centering

\begin{tabular}{|c|| c| c| c| c| c| c| c}
  & FaceTracer  & PANDA & LNets+ANet & RBM-PCA & Ours-S  & Ours-J \\
  \hline 
 Bald & 89   & 96 & 98 & 98 & 96.16 & 98.93\\ 
 \hline
 %PANDA-w &  92\%  & 61\% & 70\%  & 82\%  &  93\% & 74\%& 81\%\\
 Big Lips &  64 & 67 & 68 & 69  & 69.25 & 71.69 \\
 \hline
 %LNets+ANet(w/o) &  95\% & 66\% & 75\% & 86\%  & 94\% & 84\% &91\%\\
 Big Nose &  74  & 75 & 78 & 81  & 82.35 & 84.67\\
 \hline
 Chubby & 86   & 86 & 91 & 95  & 94.22 & 95.27 \\
 \hline
 High Cheekbones & 84   & 86 & 88 & 83  & 86.61 & 87.79 \\
 \hline
 Male & 91   & 97 & 98 & 90  & 95.65 & 98.61\\
 \hline
 Narrow Eyes & 82   & 84  & 81 & 86 & 85.45 & 87.9 \\
 \hline
 Oval Face & 64   & 65 & 66 & 73  & 74.49 & 75.94 \\
 \hline
 Young & 80   & 84 & 87 & 81  & 87.12 & 88.54 \\
 
\end{tabular}

\caption{Comparing attribute prediction models on CelebFacesA dataset.}
\label{eval_table_at}
\end{table*}

\textbf{Facial Attribute Prediction:} We leverage  identity facial attributes as an auxiliary modality for improving face identification performance. Identity facial attributes are invariant attributes which remain same from different images of a person. For example, gender, nose and lips shapes remain  the same in different images of a person; however,  attributes  such  as  glasses,  mustaches,  or  beards may or may not exist in different images of a person.  We discard such attributes in our model because we look for robust as well as invariant facial attributes. Identity facial attributes in CelebA dataset are listed as follows: narrow eyes, big nose, pointy nose, chubby, double chin, high cheekbones, male, bald, big lips and oval face . We evaluate our attribute predictor by using accuracy metric. 
\subsection{Methods for Comparisons}
\textbf{Attribute Prediction:} We  compare  our  method  with  several  competitive  algorithms including FaceTracer, PANDA\cite{zhang2014panda}, ANet+LNet \cite{liu2015deep} and MT-RBM-PCA \cite{ehrlich2016facial}. FaceTracer \cite{kumar2008facetracer}  extracts handcraft features including color histogram and HOG from some functional face image region and then concatenates these features to  train a SVM classifier for predicting attributes. Functional  regions  are determined by using ground truth landmarks. PANDA mainly was proposed by creating an ensemble of several CNNs for body attributes prediction. Each CNN in this model extracts features from a well-aligned human part using poselet. Next,  all of the extracted features are concatenated to train a SVM for body attribute prediction.  However, for our case, it is simple to adjust this method for facial attribute prediction such that the  face  part  is  aligned  using  landmark points. In ANet+LNet method, images of the first 8000 identities, which is roughly 162k images,  are  employed  for  pre-training  and  face localization. The  images  of  the  next 1000 identities,  which  is roughly 20k images, are used to train a SVM classifier. We use same testing and training sets to conduct our experiment. We compare our model with the other methods for attribute prediction. Table.\ref{eval_table_at} shows  the model improvement on identity facial attribute prediction once the model trains both tasks jointly. The results shows that joint-training has higher contribution for the attributes of  gender, bald, narrow eyes, big lip, big nose, oval face, young, high cheekbone and chubby, respectively.

\textbf{Face Identification:} We  compare  our  method  with the exiting methods on face identification which are reported from the official websites of  MegaFace\footnote{http://megaface.cs.washington.edu/results/facescrub.html}. We primarily compare with publicly released methods, for which the details are known. These methods are listed as follows:  Google FaceNet \cite{schroff2015facenet}, Center Loss \cite{wen2016discriminative}, Lightened CNN \cite{wu2015lightened}, LBP \cite{ahonen2006face} and Joint Bayes model \cite{chen2012bayesian}.

\begin{figure}
\includegraphics[scale=0.11]{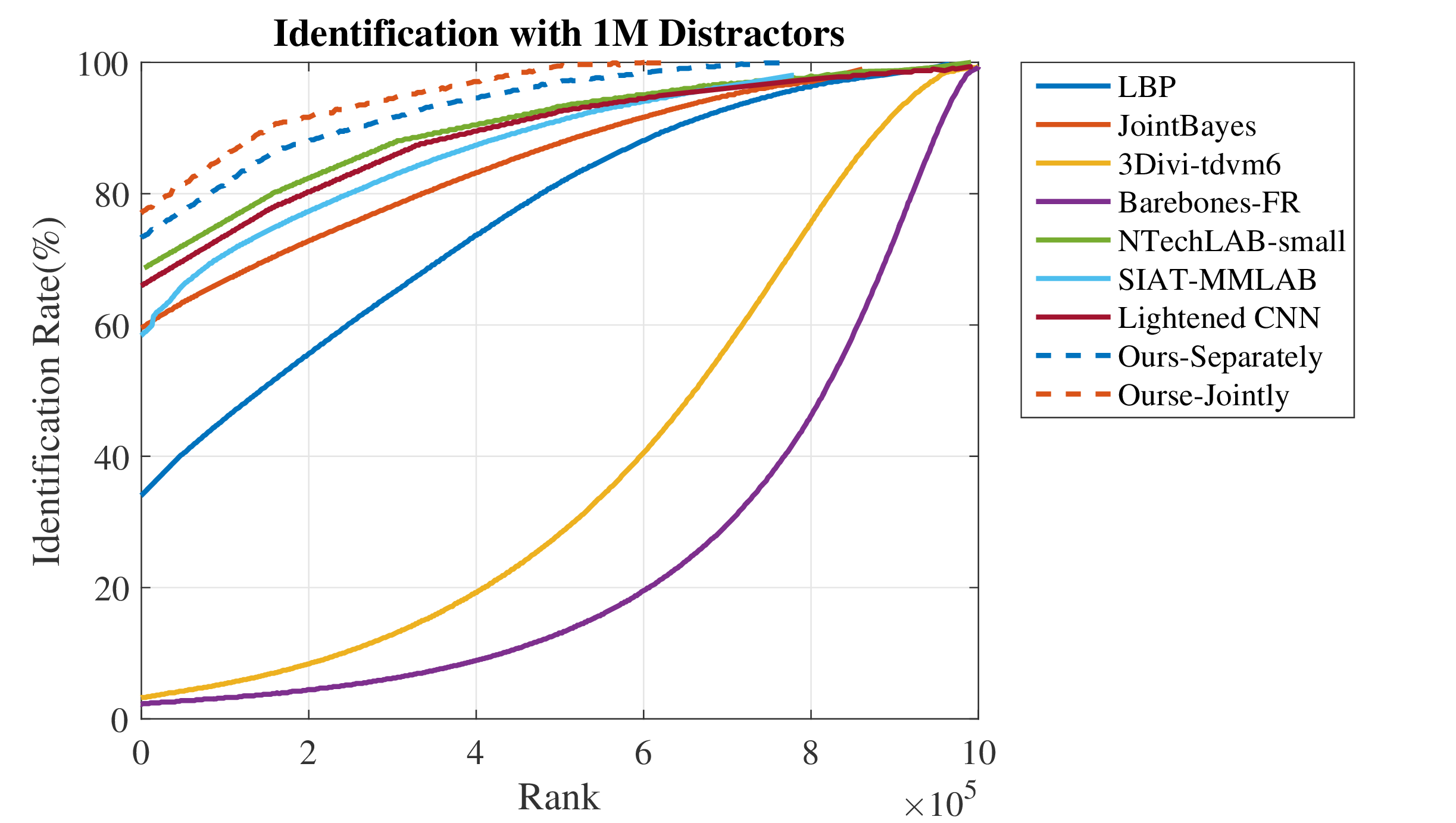}
\begin{center}
\caption{CMC curves of different methods with the protocol of small training set by 1M distractors. Please note that results of the other methods are reported from official website of MegaFace dataset.}
\label{cm}
\end{center}
\end{figure}

There are several other methods from commercial companies such as FaceAll, NTechLAB, SIAT MMLAB,
BareBonesFR, 3DiVi companies, the details of which  are  not known to the community yet. Therefore, we can not compare these methods with ours fairly; however, we report these methods to provide a comprehensive list of references on the Megaface dataset. Fig. \ref{cm} represents CMC curves for different methods; it is shown that our model covers larger area under the curve in comparison to the other methods. We also compare our model performance when the model trains facial attributes prediction and face identification jointly and separately. The results show that our face identifier benefits from joint training. We also compare performance of the algorithms by rank-1 identification accuracy; Table.\ref{eval_table} compares  face  identification  models  on  MegaFacedataset using rank-1 identification accuracy metric. The results show the superiority of our model. We also observe that the model performance increases about 2.5\% if the model train attributes and face jointly in comparison to the case which the model is trained separately.
\begin{table}
\centering
\begin{tabular}{|c||c|c|c|}
\hline
 Methods &  Rels  & Protocol & Acc\%  \\
\hline
Google - FaceNet v8   & \checkmark & Large & 70.5\\ 
%\hline 
 %PANDA-w &  92\%  & 61\% & 70\%  & 82\%  &  93\% & 74\%& 81\%\\
%\hline
 NTechLAB - Large & $\times$ & Large & 73.3\\
% \hline
 %LNets+ANet(w/o) &  95\% & 66\% & 75\% & 86\%  & 94\% & 84\% &91\%\\
% \hline
 Faceall Co. - Norm-1600 & $\times$ & Large & 64.8\\
 %\hline
Faceall Co. - FaceAll-1600 & $\times$ & Large & 63.98\\
 \hline
Lightened CNN   & \checkmark & Small & 67.11\\ 
%\hline 
 %PANDA-w &  92\%  & 61\% & 70\%  & 82\%  &  93\% & 74\%& 81\%\\
%\hline
 Center Loss & \checkmark & Small & 65.23\\
% \hline
 %LNets+ANet(w/o) &  95\% & 66\% & 75\% & 86\%  & 94\% & 84\% &91\%\\
% \hline
LBP  & \checkmark & Small & 3.02\\
 %\hline
Joint Bayes & \checkmark & Small & 2.33\\
NTechLAB -Small   & $\times$ & Small & 58.22\\ 
%\hline 
 %PANDA-w &  92\%  & 61\% & 70\%  & 82\%  &  93\% & 74\%& 81\%\\
%\hline
3DiVi Company & $\times$ & Small & 33.71\\
% \hline
 %LNets+ANet(w/o) &  95\% & 66\% & 75\% & 86\%  & 94\% & 84\% &91\%\\
% \hline
SIAT-MMLAB  & $\times$ & Small & 65.23\\
 %\hline
Barebones FR & $\times$ & Small & 59.36\\
Wang et al \cite{wang2017multi} & \checkmark & Small & 77.74\\
 \hline
 PM-Separately & \checkmark & Small & 76.15\\
 PM-Jointly & \checkmark & Small & \textbf{78.82}\\
 \hline
\end{tabular}
\caption{Comparing face identification models on MegaFace dataset using rank-1 identification accuracy metric.}
\label{eval_table}
\end{table}
\begin{figure}
\includegraphics[scale=0.45]{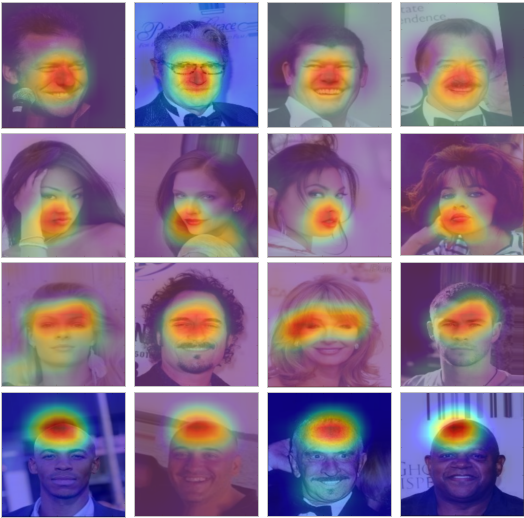}
\begin{center}
\caption{Example of class activation map generated from  attribute predictor part of our model. Each row indicates nose attribute , mouth attribute, eyes attribute  and  head attribute , respectively. We observe that highlighted regions are activated by class activation map algorithm.}
\label{camm}
\end{center}
\end{figure}
\subsection{Further Analysis}

Experimental results included in Table.\ref{eval_table} show that our model improves face recognition performance by leveraging identity facial attributes. To verify this claim, we conducted  experiments for two different cases described earlier. In the second case we emphasize predicting facial attributes, because in a real  face identification scenario, such information is not available during the testing phase. To use facial attributes as an auxiliary modality in our proposed model for face identification, we fused this modality with the last pooling layer of the model shown in Fig.\ref{fig:arc}. The second case of our model which uses the predicted attributes outperforms the first case  which does not use any privilege data.

Experimental results  show that training the two tasks jointly increases not only face identification performance, but also facial attribute prediction performance, especially on identity facial attributes such as gender. For example, experiments performed on the CelebA dataset indicate that performance on face attributes including narrow eyes, big nose, pointy nose, chubby, double chin, high cheekbones, male, bald, big lips and oval face is improved around 2\% on average if the tasks are trained jointly. Moreover, as shown in Table.\ref{eval_table}, our proposed model outperforms the accuracy of the state of the art methods for identity facial attributes prediction. One of the intuitive reasons causing this improvement is that, once our deep CNN model is trained to identify face images, it also learns more accurate face attributes in order to perform better face identification. In other words, these two modalities enhance each others' performance once they are trained jointly.

Table.\ref{eval_table} also indicates that using facial attributes as privileged data boosts the model performance on face identification task. Our model beats most of the face identification algorithms used in the  MegaFace data set challenge.

Inspired by the work in \cite{zhou2016learning}  on class activation map, we interpret the prediction decision made by our proposed architecture. Fig.\ref{camm} shows the class activation map for predicting big nose, big lips, narrow eyes and bald, respectively. We can see that our model is triggered by different semantic regions of the image for different predictions. Fig.\ref{camm} shows that our model due to using GAP layer also learns to localize the common visual patterns for the same facial attribute. Furthermore,  the deep features obtained from our attribute predictor branch can also be used for generic facial attribute localization in any given image without using any extra information such as bounding box.

%%\begin{figure}[h!]
%%\includegraphics[trim={0cm 0 0 0},scale=0.155]{LFWComparing1.eps}
%%\begin{center}
%%\caption{Face recognition performance on test data in three scenarios; GT, PA and NPD represent performance of the model when it uses ground truth, predicted attributes and no privilege data respectively.   }
%%\end{center}
%%\end{figure}
\section{Conclusion}
In this paper, we proposed an end to end  deep network to predict facial attributes and identify face images simultaneously with better performance.  Our model trains these two tasks jointly through shared CNN feature space, and also fuses predicted identity attributes modality with face  modality features  to  improve  face identification performance. The model increases both face recognition and face attribute prediction performance in comparison to the case when the model is trained separately. Experimental results show the superiority of the model in comparison to the current face identification models. The model also predicts identity facial attributes better than the state of the art models.

{\small
\bibliographystyle{ieee}
\bibliography{egbib.bib}
}
\end{document}